\documentclass[letterpaper, 10 pt, conference]{ieeeconf}  

\IEEEoverridecommandlockouts                              

\usepackage[T1]{fontenc}
\usepackage{graphicx}
\usepackage{amsmath}
\usepackage{booktabs}
\usepackage{amssymb}
\usepackage{hyperref}
\usepackage{xcolor}
\usepackage{amsmath}   
\usepackage{commath}
\usepackage[linesnumbered,ruled,vlined, noend]{algorithm2e}
\usepackage{subcaption}
\usepackage{bm}
\usepackage{wrapfig}

\SetCommentSty{mycommfont}

\DeclareMathOperator*{\argmax}{arg\,max}

\newcommand{\X}{\mathcal{X}}
\newcommand{\Y}{\mathcal{Y}}
\newcommand{\Z}{\mathcal{Z}}
\newcommand{\T}{\mathcal{T}}
\newcommand{\R}{\mathcal{R}}

\title{\LARGE \bf
Marginal MAP Estimation for Inverse RL under Occlusion with Observer Noise
}

\author{Prasanth Sengadu Suresh$^{1}$ and Prashant Doshi$^{2}$
\thanks{$^{1}$Prasanth Sengadu Suresh is with THINC Lab, Computer Science,
        University of Georgia, Athens, GA
        {\tt\small ps32611@uga.edu}}%
\thanks{$^{2}$Prashant Doshi is with the Department of of Computer Science and Institute for Artificial Intelligence, University of Georgia,
        Athens, GA
        {\tt\small pdoshi@uga.edu}}%
}

\begin{document}

\maketitle
\thispagestyle{empty}
\pagestyle{empty}

\begin{abstract}

We consider the problem of learning the behavioral preferences of an expert engaged in a task from noisy and partially-observable demonstrations. This is motivated by real-world applications such as a line robot learning from observing a human worker, where some observations are occluded by environmental objects that cannot be removed. Furthermore, robotic perception tends to be imperfect and noisy. Previous techniques for inverse reinforcement learning (IRL) take the approach of either omitting the missing portions or inferring it as part of expectation-maximization, which tends to be slow and prone to local optima. We present a new method that generalizes the well-known Bayesian maximum-a-posteriori (MAP) IRL method by marginalizing the occluded portions of the trajectory. This is additionally extended with an observation model to account for perception noise. We show that the marginal MAP (MMAP) approach significantly improves on the previous IRL technique under occlusion in both formative evaluations on a toy problem and in a summative evaluation on an onion sorting line task by a robot.

\end{abstract}

\section{Introduction}

Inverse reinforcement learning (IRL) aims to infer an expert's behavioral preferences from observations of the expert performing the task as a way to learn the expert's behavior. It represents an important paradigm in the toolkits for robot learning from demonstrations~\cite{argall2009survey} and imitation learning~\cite{hussein2017imitation}. To perform this inference feasibly, the typical IRL methodology is to ascribe a decision-making model to the expert, which it solves optimally and follows the obtained policy~\cite{ng2000algorithms,abbeel2004apprenticeship}. Subsequently, IRL methods learn the reward function of the decision-making model that best explains the observed behavior under the assumption that the other components of the model are known (recent techniques relax this assumption -- see Section 6.3 of the survey~\cite{Arora21:Survey} for a review of such methods).
    Another assumption prevalent among IRL methods, which is particularly impractical for robot learning, is that the expert's behavior is observed fully and perfectly. Recent work~\cite{Bogert_thesis} has partially relaxed this assumption by recognizing that portions of the observed trajectories may be occluded from the learner's view in real-world applications. However, scarce attention has been given to the associated challenge that a robotic observer often has a noisy sensor model or a noisy perception pipeline. As such, the learner's perception of the task performance may be both incomplete and imperfect~\cite{Arora21:Survey}. For example, consider a line robot tasked with learning how to sort vegetables such as onions by watching the human perform the sort. Positioning an unobtrusive depth-camera to avoid occlusion from other nearby workers on the line is challenging and correctly discriminating between blemished and unblemished onions is not always possible.
    We present the first IRL method that allows learning from trajectories, which contain both occlusions and the result of noisy perception. We adopt Bayesian IRL~\cite{Ramachandran2007} as our point of departure and generalize the maximum-a-posteriori (MAP) inference~\cite{choi2011map} of the reward function in two ways:  
    \begin{itemize}
    \item First, we introduce a noisy observation model in the MAP inference framework under the assumption that the robot's observation model is known.
    \item Second, given that the portions of the trajectory suffering from occlusion are known (a realistic assumption as such portions are easily detected), we may model the observed trajectory as the full trajectory with the occluded elements  marginalized. Consequently, we perform a marginalized MAP inference (MMAP) of the reward function.        
    \end{itemize}

A forward-backward search of the hidden variable values yields a probable list of observations that makes the marginalization more efficient than simply summing over all possible observations. We evaluate our MMAP-BIRL on a toy problem and on the use-inspired domain of learning to sort onions. We show that it improves dramatically on a previous method that uses expectation-maximization for occlusion~\cite{Bogert_EM_hiddendata_fruit} in both learning accuracy and run time. Our experiments with the physical cobot Sawyer demonstrate that the learned policy allows Sawyer to sort onions on a conveyor with both improved precision and recall. 

\section{Preliminaries: Bayesian IRL and MAP Inference}
\label{sec:background}

IRL utilizes input from an expert whose behavior is assumed to be modeled using a Markov decision process (MDP)~\cite{Puterman1994}, which it solves optimally. The expert provides demonstrations of the task to the learner and the problem is to solve for the expert's reward function that best explains the observed behavior. Formally, the MDP of an expert is defined as a quadruple $\langle$$S$, $A$, $T$, $R$$\rangle$, where $S$ is the set of  states defining the environment, $A$ is the expert's set of possible actions, $T: S \times A \times S \rightarrow [0,1]$ gives the transition probabilities from any given state to a next state for each action, and $R: S \times A \rightarrow \mathbb{R}$ is the reward function modeling the expert's preferences, rewards, or costs of performing an action from a state. Typically, the learner is aware of the expert's MDP except for the reward function.

We may model the reward function as a linearly-weighted sum of $K$ basis functions~\cite{ng2000algorithms}: $R_{\bm{\theta}}(s,a)$ $\triangleq \sum_{k=1}^K \theta_k\phi_k(s,a)$ where $K$ is finite and non-zero, $\theta_k$ are the weights, and $\phi: (S,A) \rightarrow (0,1)$ is a feature function. A binary feature function maps a state from the set of states $S$ and an action from the set of actions $A$ to 0 (false) or 1 (true). 
A (stationary) \textit{policy} is a mapping from states to actions $\pi: S \rightarrow A$ and the discounted, infinite-horizon value of a policy $\pi$ for a given reward function $R_{\bm{\theta}}$ at some state $s \in S$, with $t$ denoting time steps is given by:
$E_{s}\left[V^\pi(s)\right] = E \left[\sum_{t=0}^\infty \gamma^t R_{\bm{\theta}} (s^t,\pi(s^t))| \pi, s\right].$
In this paper, we consider the situation where a portion of the trajectory is occluded from the learner. In keeping with previously established notation~\cite{Bogert_EM_hiddendata_fruit}, let the set of input trajectories of finite length $\mathcal{T}$ generated by an MDP attributed to the expert be, $\mathcal{X}^\mathcal{T} = \{X|X = Y \bigcup Z \}$.
Here, $Y$ is the observable portion and $Z$ is the occluded part of a trajectory $X$. The complete trajectory $X$ is a sequence, $X$ = $(s^1, a^1, s^2, a^2, s^3, ..., s^{\T}, a^{\T})$; some of these may be occluded. We build on the well-known Bayesian approach to IRL (BIRL)~\cite{ramachandran2007bayesian} that treats the reward function as a random variable and utilizes a prior distribution over the reward function, given as 
\begin{small}
\begin{align}
    P(R) = \prod\nolimits_{s \in S, a \epsilon A} Pr(R(s,a)).
    \label{eqn:prior}
\end{align}
\end{small}
Notice that the reward values for the state-action pairs are i.i.d. Ramchandran and Amir~\cite{ramachandran2007bayesian} discuss some example prior distributions including the Gaussian. We may derive the likelihood function for the demonstrated set of trajectories $\X$ as: 
\begin{small}
\begin{align}
    &P(\X|R) = \prod_{X=1}^{\X}\prod_{t=1}^{\T} Pr(s^t_X,a^t_X;R)=\prod_{X=1}^{\X} Pr(s^1_X)Pr(a^1_X|s^1_X;R)\nonumber\\
&\times~\prod_{t=1}^{\T-1} Pr(s^{t+1}_X|s^t_X,a^t_X)~Pr(a^{t+1}_X|s^{t+1}_X;R) = \prod\nolimits_{X=1}^{\X}Pr(s^1_X)\nonumber\\
&\times~\pi(a^1_X|s^1_X;R)~\prod\nolimits_{t=1}^{\T-1} ~T(s^t_X,a^t_X,s^{t+1}_X)~\pi(a^{t+1}_X|s^{t+1}_X;R).
\label{eqn:likelihood}
\end{align}
\end{small}
The policy is commonly modeled in BIRL as a Boltzmann exploration~\cite{ramachandran2007bayesian,vroman2014maximum} of the form: 
\begin{align}
\pi(a|s;R) = \frac{e^{\beta~Q(s,a;R)}}{\sum_{a' \in A} e^{\beta~Q(s,a';R)}} = \frac{e^{\beta~Q(s,a;R)}}{\Xi(s)}
\label{eqn:policy}
\end{align}
where $\Xi(s)$ is the partition function. As the Boltzmann temperature parameter $\beta$ becomes large, the exploration assigns increasing probability to the action(s) with the largest Q-value(s). $\beta$ ranges between 0-1 with 0 being fully exploratory and 1 being fully greedy. Methods for both maximum likelihood~\cite{vroman2014maximum,Jain19:Model} and maximum-a-posteriori~\cite{choi2011map} inferences of the reward function exist, which use the likelihood function of Eq.~\ref{eqn:likelihood} and, in case of MAP inference, the prior as well. MAP inference for IRL has been shown to be more accurate, benefiting from its use of the prior~\cite{choi2011map}.  Formally, we may write MAP inference in log form as:
\begin{small}
\begin{align}
        R^*=\argmax_{\R} Pr(R|\X)\nonumber=\argmax_{\R} ~\log Pr(\X|R)+\log Pr(R) 
\end{align}
\end{small}
where $\R$ is the continuous space of reward functions, and the prior and likelihood functions are as given in Eqs.~\ref{eqn:prior} and~\ref{eqn:likelihood}, respectively.

Choi et al.~\cite{choi2011map} presents a gradient-based approach to obtain $R^*$, which searches the reward optimality region only. Given the expert's policy, Ng and Russell~\cite{ng2000algorithms} show that this region can be obtained as:
\begin{small}
\begin{align}\label{R_opt_rgn}
    H^\pi = I - (I^A - \gamma T)(I - \gamma T^\pi)^{-1} E^\pi 
\end{align}
\end{small}
where $I$ is the identity matrix, $T$ is the transition matrix, $E^\pi$ is an $|S|$ $\times$ $|S|$ $\times$ $|A|$ matrix with the $(s, (s', a'))$ element being 1 if s = $s'$ and $\pi(s')$ = $a'$. $I^A$ is an $|S| \times |A|$ $\times$ $|S|$ matrix constructed by stacking the $|S|$ $\times$ $|S|$ identity matrices $|A|$ times. The reward update rule is given as $R_{new} \leftarrow R + \delta_t \nabla_{R} Pr(R|\X)$\label{update_rule} where $\delta_t$ is an appropriate step size (or the learning rate). As computing $\nabla_{R} Pr(R|\X)$ involves calculating an optimal policy, this may slow down the computations. By checking if the gradient lies within the new reward optimality region, we can reuse the same gradient and reduce the computational time.  If $H^\pi . R_{new} \leq 0$, then the previous gradient is reusable.

\section{IRL under Occlusion with Noisy Observations}
\label{sec:mmap-birl}

The traditional MAP-BIRL assumes that the input trajectories are noiseless and fully-observable. However, these assumptions are difficult to satisfy in real-world use cases of robotic learning. In particular, it may be difficult to position a depth-camera in a factory such that the processing line task is fully observed. Furthermore, recording sensors as well as the visual state-action recognition tend to be noisy. 

Consequently, we generalize MAP-BIRL to learn in the context of both occlusions and noisy learner observations.
Let $X$ = $(o^1_l, o^2_l, o^3_l, ..., o^{\T}_l)$ where each element $o^t_l$ is the learner's observation of the expert at a time step $t$; some of these observations may be occluded. We begin by using the parameterized linear sum of reward weights, $R_{\bm{\theta}}$, as the representation of the reward function. Subsequently, the prior over the reward function (Eq.~\ref{eqn:prior}) is now a distribution over the feature weights, decomposed into independent distribution over each weight:
\begin{small}
\begin{align}
    Pr(R_{\bm{\theta}}) = \prod\nolimits_{\theta \in \Theta} Pr(\theta)
    \label{eqn:prior-weights}
\end{align}
\end{small}

\vspace{-0.15in}
\subsection{Framework}
Obviously, we may simply ignore the occluded data and utilize just the observed portions of the set of trajectories $\Y{}$ for IRL~\cite{Bogert_mIRL_Int_2014}. In other words, $R_{\bm{\theta}}^* = \argmax_{\bm{\theta} \in \bm{\Theta}} Pr(R_{\bm{\theta}}|\Y)$
     $= \argmax_{\bm{\theta} \in \bm{\Theta}} Pr(\Y | R_{\bm{\theta}})~Pr(R_{\bm{\theta}})$. But, as Bogert et al.~\cite{Bogert_EM_hiddendata_fruit} shows, IRL's performance improves if the occluded portion can be inferred because it may contain salient state-action pairs. As such, we formulate the {\em marginal MAP} inference of the reward function from the data. To enable this, the likelihood of the visible portions of the trajectories can be written as the marginal of the complete trajectory $X$ by summing out the corresponding hidden portion $Z$:
\begin{small}
\begin{align*}
    &Pr(\Y| R_{\bm{\theta}}) = \prod\nolimits_{Y \in \Y} Pr(Y|R_{\bm{\theta}}) = \prod\nolimits_{Y \in \Y} \sum\nolimits_{Z \in \Z} Pr(Y,Z| R_{\bm{\theta}}) \nonumber\\
    &= \prod\nolimits_{Y \in \Y} \sum\nolimits_{Z \in \Z}~Pr(X | R_{\bm{\theta}}).
\end{align*}
\end{small}
Here, the parameters $\bm{\theta}$ are the maximization variables and the occluded portion $Z$ of a trajectory comprises the summation variables of the marginal MAP inference. Using the above likelihood function, the MMAP-BIRL problem is more specifically formulated as:
\begin{small}
\begin{align}
    R_{\bm{\theta}}^* = \argmax_{\bm{\theta} \in \bm{\Theta}} ~\prod\nolimits_{Y \in \Y} \sum\nolimits_{Z \in \Z} Pr(Y,Z | R_{\bm{\theta}})~Pr(R_{\bm{\theta}}).
     \label{eq:mmap-birl}  
\end{align}
\end{small}
Let $Z$ be the collection of the  observations in the occluded time steps of $X$, and $Y = X/Z$. Then,  
\begin{small}
\begin{align*}
R_{\bm{\theta}}^* = \argmax_{\bm{\theta} \in \bm{\Theta}}\prod\nolimits_{Y \in \Y} \sum\nolimits_{Z \in \Z} Pr(o^1_l, o^2_l, o^3_l, ..., o^\T_l| R_{\bm{\theta}})Pr(R_{\bm{\theta}}).
\end{align*}
\end{small}
The learner's observation $o^t_l$ is a noisy perception of the expert's state and action at time step $t$, and the observations are conditionally independent of each other given the expert's state and action. Therefore, we introduce the state-action pairs in the likelihood 
function above.  
\begin{small}
\begin{align*}
 &Pr(o^1_l, o^2_l, o^3_l, \ldots, o^{\T}_l| R_{\bm{\theta}}) 
 = \nonumber\\
 &\sum\limits_{s^1, a^1, ..., s^{\T}, a^{\T}}Pr(o^1_l,o^2_l,o^3_l,...,o^{\T}_l, s^1, a^1, s^2, a^2, ..., s^{\T}, a^{\T}| R_{\bm{\theta}}).
\end{align*}
\end{small}
For convenience, let $\tau$ denote the underlying trajectory of state-action pairs, $\tau = (s^1, a^1, s^2, a^2...,s^{\T}, a^{\T})$. Then, we may reformulate the MMAP-BIRL problem as:
\begin{small}
\begin{align}
&R_{\bm{\theta}}^* = \argmax_{R_{\bm{\theta}}}\prod\nolimits_{Y \in \Y} \sum\nolimits_{Z \in \Z} \sum\nolimits_{\tau \in (|S||A|)^{\T}} Pr(o^1_l, o^2_l, o^3_l, \nonumber\\
&\ldots, o^\T_l, \tau| R_{\bm{\theta}})~Pr(R_{\bm{\theta}}).
\end{align}
\end{small}

\vspace{-0.15in}
\subsection{MMAP-BIRL Gradients}

The MMAP inference problem is hard. Previous approaches, mostly in the context of Bayesian network inference, have utilized AND-OR graph structures to perform the inference~\cite{marinescu2014and}. But, the maximization variables in these techniques are discrete, which allows the use of a discrete data structure such as a graph to model the inference. As our maximization variables are continuous, we seek to solve the hard MMAP inference problem using gradient ascent techniques.

The log forms of the prior and the likelihood function are represented respectively as:
\begin{small}
\begin{align*}
    &L_{\bm{\theta}}^{pr} = \log Pr(R_{\bm{\theta}}) \mbox{ and } \nonumber\\
    &L_{\bm{\theta}}^{lh} = \nonumber\\
    &\sum\nolimits_{Y \in \Y} \log\sum\nolimits_{Z \in \Z} \sum\nolimits_{\tau \in (|S||A|)^{\T}}Pr(o^1_l, o^2_l, o^3_l, \ldots, o^\T_l, \tau| R_{\bm{\theta}}).
\end{align*}
\end{small}
If we let the prior $Pr(\theta)$ in Eq.~\ref{eqn:prior-weights} be Gaussian (some values of each feature weight are more likely than others), i.e., 
$Pr(\theta; \mu_\theta, \sigma_\theta) = \frac{1}{\sqrt{2\pi}\sigma_\theta}e^{-\frac{(\theta - \mu_\theta)^2}{2\sigma_{\theta}^2}}$, where the mean $\mu_\theta$ and standard deviation $\sigma_\theta$ may differ between the feature weights. Then, the gradient of the log prior is easily obtained as, 
\begin{align}
   \frac{\partial L_{\bm{\theta}}^{pr}}{\partial \theta} = \frac{- (\theta - \mu_\theta)}{2 \sigma_\theta^2}.
\label{eqn:gradient_prior}
\end{align}

Next, to obtain the gradient of $L_{\bm{\theta}}^{lh}$, we expand the term $Pr(o^1_l, o^2_l, o^3_l, \ldots, o^\T_l, \tau| R_{\bm{\theta}})$. 
\begin{small}
\begin{align*}
&Pr(o^1_l, o^2_l, o^3_l, \ldots, o^\T_l, \tau| R_{\bm{\theta}}) \\
&= Pr(o^1_l, o^2_l, o^3_l, \ldots, o^\T_l|\tau, R_{\bm{\theta}})~Pr(\tau|R_{\bm{\theta}})\\
&= Pr(o^1_l|o^2_l, o^3_l, ..., o^\T_l,\tau,R_{\bm{\theta}})~Pr(o^2_l, o^3_l, ..., o^\T_l,\tau|R_{\bm{\theta}})~Pr(\tau|R_{\bm{\theta}})\\ 
&= Pr(o^1_l|s^1, a^1)~Pr(o^2_l, o^3_l, \ldots, o^\T_l,\tau'|R_{\bm{\theta}})~Pr(\tau|R_{\bm{\theta}})
\end{align*}
\end{small}
We obtain the last step by noting that the learner's current observation is conditionally independent of its future observations given the expert's true state and action. Let $O_l(s^1,a^1,o^1_l)$ represent $Pr(o^1_l|s^1, a^1)$, which is the learner's stochastic mapping from the expert's state and performed action $(s^1, a^1)$ to the learner's observation of it, $o^1_l$, all corresponding to the same time step. It informs the learner about the expert's state and performed action, albeit noisily.   

Using the above observation model, we may continue expanding $Pr(o^2_l, o^3_l, \ldots, o^\T_l, \tau'| R_{\bm{\theta}})$ as,
\begin{small}
\begin{align*}
&Pr(o^1_l, o^2_l, o^3, \ldots, o^\T_l, \tau| R_{\bm{\theta}}) = O_l(s^1, a^1, o^1_l)Pr(o^2_l, o^3_l, \nonumber\\
&\ldots, o^\T_l,\tau'|R_{\bm{\theta}})Pr(\tau|R_{\bm{\theta}})= \prod\nolimits_{t=1}^\T~O_l(s^t, a^t, o^t_l)~Pr(\tau|R_{\bm{\theta}}).
\end{align*}
\end{small}
Observe that the last term $Pr(\tau|R_{\bm{\theta}})$ corresponds exactly to $Pr(X|R_{\bm{\theta}})$ in Eq.~\ref{eqn:likelihood} for a trajectory of state-action pairs $X \in \X$. Hence, we substitute the terms inside the outer product of Eq.~\ref{eqn:likelihood} above,
\begin{small}
\begin{align*}
&Pr(o^1_l, o^2_l, o^3_l, \ldots, o^\T_l, \tau| R_{\bm{\theta}}) = \prod\nolimits_{t=1}^\T O_l(s^t, a^t, o^t_l)Pr(s^1)\nonumber\\
&\times\pi(a^1|s^1;\bm{\theta})\prod_{t'=1}^{\T-1} T(s^{t'},a^{t'},s^{t'+1})\pi(a^{t'+1}|s^{t'+1};\bm{\theta}) = Pr(s^1) \times\nonumber\\
&\pi(a^1|s^1;\bm{\theta})\left (\prod\nolimits_{t=1}^{\T-1} O_l(s^t, a^t, o^t_l)~T(s^t,a^t,s^{t+1})~\pi(a^{t+1}|s^{t+1};\bm{\theta}) \right )\nonumber\\
&\times~O_l(s^\T, a^\T, o^\T_l). 
\end{align*}
\end{small}
We may now rewrite the log likelihood $L_{\bm{\theta}}^{lh}$ more fully as,

\begin{small}
\begin{align}
&L_{\bm{\theta}}^{lh} = \sum\limits_{Y \in \Y} \log~\sum\limits_{Z \in \Z} \sum\limits_{\tau \in (|S||A|)^{\T}} Pr(s^1)~\pi(a^1|s^1;\bm{\theta})~\times\nonumber\\
&\left (\prod_{t=1}^{\T-1} O_l(s^t, a^t, o^t_l)T(s^t,a^t,s^{t+1})\pi(a^{t+1}|s^{t+1};\bm{\theta}) \right )O_l(s^\T, a^\T, o^\T_l).
\label{eqn:log_likelihood_full}
\end{align} 
\end{small}
While obtaining the gradient of $L_{\bm{\theta}}^{lh}$ is not trivial, it is possible and we show it below. The complete derivation of this gradient is given in the supplementary file.
\begin{small}
\begin{align*}
&\frac{\partial L_{\bm{\theta}}^{lh}}{\partial \bm{\theta}} = \sum\limits_{Y \in \Y}\left (\sum\limits_{Z \in \Z} \sum\limits_{\tau \in (|S||A|)^{\T}} Pr(s^1)~\pi(a^1|s^1;\bm{\theta}) \right. ~\times \nonumber\\ 
&\left( \prod_{t=1}^{\mathcal{T}-1} O_l(s^t,a^t,o^t_l)T(s^{t},a^{t},s^{t+1})\pi(a^{t+1}|s^{t+1};\bm{\theta}) \right. \left . O_l(s^\T, a^\T, o^\T_l)\right )^{-1} \nonumber\\ 
&\sum\limits_{Z \in \Z} \sum\limits_{\tau \in (|S||A|)^{\T}} Pr(s^1)\pi(a^1|s^1;\bm{\theta}) \left( \prod_{t=1}^{\mathcal{T}-1}  O_l(s^t,a^t,o^t_l)T(s^{t},a^{t},s^{t+1}) \right . \nonumber\\ 
& \times \left . \left (\sum\limits_{t=1}^{\mathcal{T}-1} \frac{\partial \pi(a^{t+1}|s^{t+1};\bm{\theta})}{\partial\theta} \prod\limits_{k \neq t}^{\mathcal{T}-1} \pi(a^k|s^k;\bm{\theta})\right )\right )~O_l(s^\T, a^\T, o^\T_l).
\label{eqn:gradient_likelihood} 
\end{align*}
\end{small}
Partially differentiating the policy whose form is shown in Eq.~\ref{eqn:policy}, we get,
\begin{small}
\begin{align*}
    &\frac{\partial \pi(a^{t+1}|s^{t+1};\bm{\theta})}{\partial\bm{\theta}} = \pi(a^{t+1}|s^{t+1};\bm{\theta})(\dfrac{\beta~\partial Q^*(s^{t+1}, a^{t+1}; \bm{\theta})}{\partial \bm{\theta}} ~- \nonumber\\
    &\sum_{a'\in A} \pi(a'|s^{t+1};\bm{\theta})\dfrac{\beta~\partial Q^*(s^{t+1},a'; \bm{\theta})}{\partial \bm{\theta}})\nonumber
\end{align*}
\end{small}
where the partial derivative of the $Q$-function can be obtained as:
\begin{small}
\begin{align*}
    &\dfrac{\partial Q^*(s^{t+1},a^{t+1}; \bm{\theta})}{\partial \bm{\theta}} = \dfrac{\partial R_\theta(s^{t+1}, a^{t+1})}{\partial \bm{\theta}} ~+ \nonumber\\
    &\gamma \sum_{s' \in S}T(s^{t+1}, a^{t+1},s')\sum_{a'\in A} \pi(a'|s^{t+1};\bm{\theta})\dfrac{\partial Q^*(s', a'; \bm{\theta})}{\partial \bm{\theta}}).
\end{align*}
\end{small}
The differential of $R_{\bm{\theta}}$ becomes the feature weights. As shown by Choi et al.~\cite{choi2011map}, both $Q^*$ and $V^*$ are Lipschitz continuous, convex, and differentiable based on the reward optimality condition~\eqref{R_opt_rgn}. 

\subsection{Algorithm for MMAP-BIRL using Gradient Ascent}

The algorithm for MMAP-BIRL is shown in Algorithm~\ref{alg:main}. It computes the initial gradient $\nabla_{\bm{\theta}} P(R_{\bm{\theta}}|\Y)$ for the initially sampled weights (line 1), performs forward rollout (line 7) to find the policy corresponding to these weights, computes the reward optimality region (line 8) and stores all of these (line 9). Notice that the gradient requires inferring possible $Z$, a set considerably narrowed down by forward-backward search. Then, we repeatedly update the reward weights according to the update rule. If the weights fail to satisfy the optimality condition, we find a new gradient and proceed with the remaining steps. On convergence, we return the learned weights (line 21).

\begin{algorithm}[ht!] 
\caption{MMAP-BIRL}
\label{alg:main}
\begin{footnotesize}
\SetKwInput{KwInput}{Input}                
\SetKwInput{KwOutput}{Output}              
\DontPrintSemicolon
  
  \KwInput{MDP, $\Y$, step-size $\delta_n$, $\epsilon$}
  \KwOutput{Learned reward function $R_{\bm{\theta}}$}
        Sample $R_{\bm{\theta}}$ from the prior distribution 
        
        Initialize $\Pi$ $\leftarrow$ {$\varnothing$}, $\delta$ $\leftarrow$ $\infty$ 
        
        \If{$\nabla_{\bm{\theta}}Pr(R_{\bm{\theta}}|\Y)$ not in $\Pi$}{
            \If{no. of occlusions $>$ 0}{
                $\Z \leftarrow$ Bidirectional\_Search(MDP, $\Y$)
                
                $\nabla_{\bm{\theta}} Pr(R_{\bm{\theta}}|\Y) \leftarrow$ Compute\_MMAP\_Gradient(MDP, $\Y$, $\Z$)
            }
        }
 \SetAlgoNoLine        $\pi \leftarrow$ Solve\_MDP($R_{\bm{\theta}}$)
        
         $H^{\pi} \leftarrow$ Compute\_Reward\_Optimality\_Region($\pi$) as shown in Eq.~\ref{R_opt_rgn}
        
        $\Pi \leftarrow \{\langle \pi,H^{\pi}, \nabla_{\bm{\theta}}P(R_{\bm{\theta}}|\Y) \rangle\}$

         \While{$\delta > \epsilon (1-\gamma) / \gamma$}{
        
         $R_{\bm{\theta}new}$ $\leftarrow$ $R_{\bm{\theta}}$ $+$ $\delta_n \nabla_{\bm{\theta}}Pr(R_{\bm{\theta}}|\Y)$
        
         Repeat steps 7 and 8 using $R_{\bm{\theta},new}$
        
         \If{$R_{\bm{\theta}_{new}}$ is not in the reward optimality region $H^{\pi}$}
         {
        
         Repeat steps 3 to 6 using $R_{\bm{\theta},new}$
        
       \If{isNewEntry($\langle \pi,H^{\pi}, \nabla_{\bm{\theta}}P(R_{\bm{\theta}_{new}}|\Y)\rangle$)}{
        
         Repeat step 9
         }
         }
         \Else{ReuseCachedGradient($\Pi$)}
        
         $\delta \leftarrow  \abs{ R_{\bm{\theta}} - R_{\bm{\theta}_{new}} }$
        
         $R_{\bm{\theta}}$ $\leftarrow$ $R_{\bm{\theta}_{new}}$
         }
    \Return $R_{\bm{\theta}}$ 
\end{footnotesize}
\end{algorithm}
\vspace{-0.15in}
\section{Experiments}
\label{sec:experiments}
We evaluated MMAP-BIRL on two domains. For both these domains, we use the Boltzmann temperature $\beta = 0.03$, step size $\delta_n = 0.01$, discount factor $\gamma = 0.99$, and a decay rate of 0.95. We solve the MDP in Algorithm~\ref{alg:main} using policy iteration to obtain the current iteration's policy, $Q$- and $V$- values.  
Our code for this algorithm will be made publicly available on GitHub upon publication. We evaluate the performance of MMAP-BIRL using the well-known metric of inverse learning error (ILE) and run time. ILE is inversely proportional to the accuracy of the learned reward function, ILE $= \sum_{s \in S} \lVert V^{\pi_E}(s) - V^{\pi_L}(s) \rVert$, where $V^{\pi_E}$ is the value of the expert's policy $\pi_E$ and $V^{\pi_L}$ is the value of the learned policy $\pi_L$ using the true MDP. 

We compare MMAP-BIRL's performance with an extension of Bogert et al.'s HiddenDataEM~\cite{Bogert17:Scaling} that utilizes expectation-maximization for managing the hidden portions of the trajectory. However, as the method does not account for observer noise, we generalize it by introducing the observation model $O_l$.  

\vspace{-0.05in}
\subsection{Forestworld}
\vspace{-0.05in}

\begin{figure}[ht!]
  \centering
\begin{subfigure}{.21\textwidth}
  \centering
  \includegraphics[width=1\linewidth]{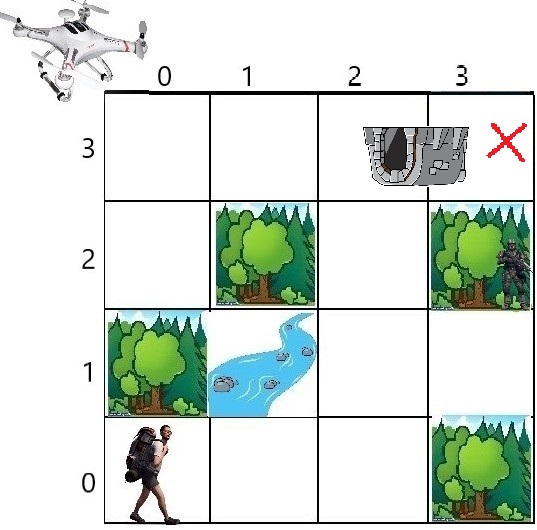}  
  \caption{}
  \label{fig:forestworld}
\end{subfigure}
\begin{subfigure}{.26\textwidth}
  \includegraphics[width=1\linewidth]{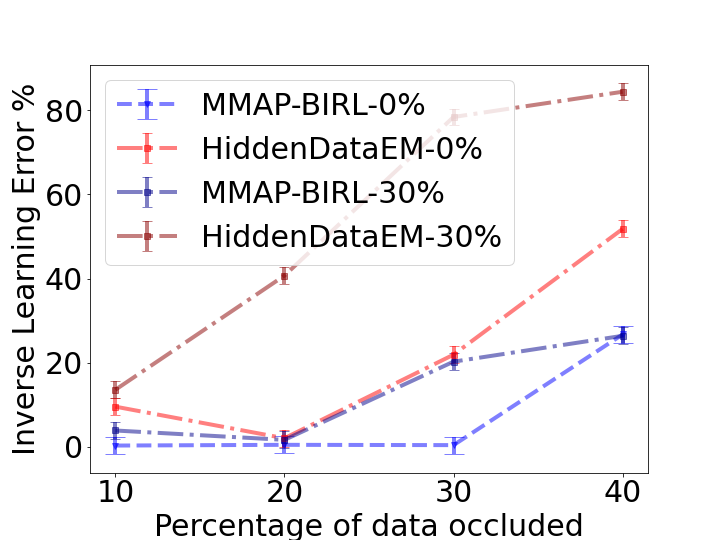}  
  \caption{}
  \label{fig:sub-second}
\end{subfigure}
\begin{subfigure}{.235\textwidth}
  \centering
  \includegraphics[width=1\linewidth]{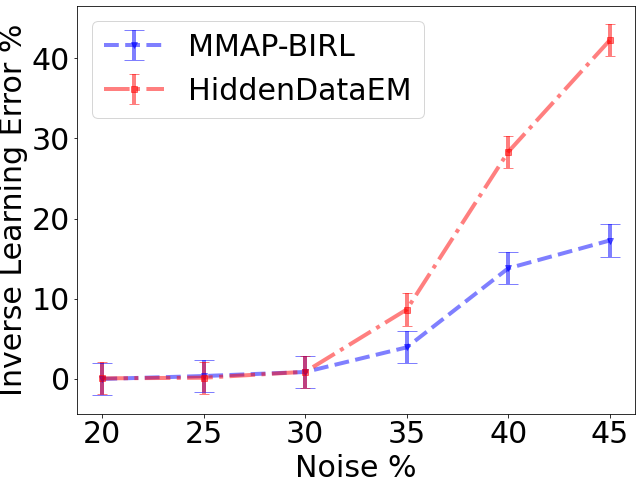}  
  \caption{}
  \label{fig:sub-third}
\end{subfigure}
\begin{subfigure}{.235\textwidth}
  \centering
  \includegraphics[width=1\linewidth]{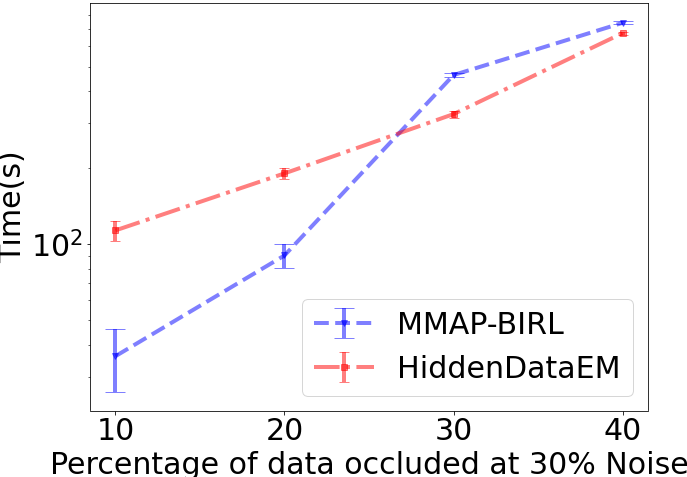}  
  \caption{}
  \label{fig:sub-forth}
\end{subfigure}
\caption{\small (a) A fugitive intends to reach the safe sector (3,3) while avoiding the river in (1,1) and the army personnel in (3,2). These sector preferences are not known to the UAV flying overhead. (b) ILE increases with increasing occlusions on noise-free data and data with 30\% noise but less so for MMAP-BIRL. (c) ILE changes with increasing noise on occlusion-free data. (d) Average clock times for increasing occlusion at 30\% noise. These  were measured on a Ubuntu PC with quad-core Xeon CPU @ 3.2GHz and 76GB RAM.}
\label{fig:forestworld-results}
\vspace{-0.2in}	
\end{figure}

Our first domain for formative evaluations is a previously introduced toy problem~\cite{Bogert_EM_hiddendata_fruit} consisting of a 4x4 grid traversed by a fugitive. A UAV is tasked with reconnaissance of the fugitive (to learn which location is the goal and which locations are avoided), but the latter's movement is not always visible due to forest cover in some sectors, as shown in Fig.~\ref{fig:forestworld}. We model the fugitive's navigation through the grid as an MDP. The states of the MDP are the sector coordinates $(x, y)$ and there are four actions corresponding to movement in the 4 cardinal directions. The start state of the fugitive may vary. 
However, the resulting next sector location is not deterministic: there is a 10\% chance that the fugitive may end up in any of the three sectors other than the intended one.   
A hidden tunnel from (2,3) to (3,3) introduces ambiguity whether the fugitive has reached the goal location of (3,3): with a chance of 30\% the UAV's sensors may incorrectly place the fugitive back at (2,3).  This forms the UAV camera's observation model. The UAV models the fugitive's reward function as a weighted linear sum of the following feature functions:
\begin{itemize}[leftmargin=*, topsep=-0.02in, itemsep=-.02in]
    \item \textsf{Avoidable\_state$(x,y)$} is activated if the fugitive eschews $(x,y)$,
    \item \textsf{Goal\_state$(x,y)$} is activated if $(x,y)$ is the fugitive's goal location.
\end{itemize}
For Fig.~\ref{fig:forestworld}, the fugitive's own reward function places a high negative weight on {\sf Avoidable\_state$(1,1)$} and {\sf Avoidable\_state$(3,2)$}, whereas a high positive weight for {\sf Goal\_state$(3,3)$}.  

We evaluate the performance of the methods under varying levels of occlusion (from 10\% to 40\%) while keeping observation noise fixed at 0\% and 30\% (Fig.~\ref{fig:sub-second}), and for varying levels of noise (from 20\% to 45\%) without occlusion (Fig.~\ref{fig:sub-third}). Each data point is the mean of 10 batches with 10 trajectories in each batch exhibiting the corresponding level of occlusion and noise, and a Gaussian prior ($\mu=-1$, $\sigma^2=.5$) is used for MMAP-BIRL. Notice that, as we may expect, the ILE increases as the learning becomes more challenging. Between the two methods, MMAP-BIRL exhibits a much lower ILE, and it does not increase as dramatically as HiddenDataEM, especially for the 30\% noise level. However, the HiddenDataEM does run marginally faster than MMAP-BIRL in this toy problem, and both show run times that generally increase {\em linearly} as the occlusion rate increases.   

\subsection{Robotic Sorting on Processing Lines}

\begin{figure}[t!]
\centering
\begin{subfigure}{.235\textwidth}
  \includegraphics[width= 1\linewidth]{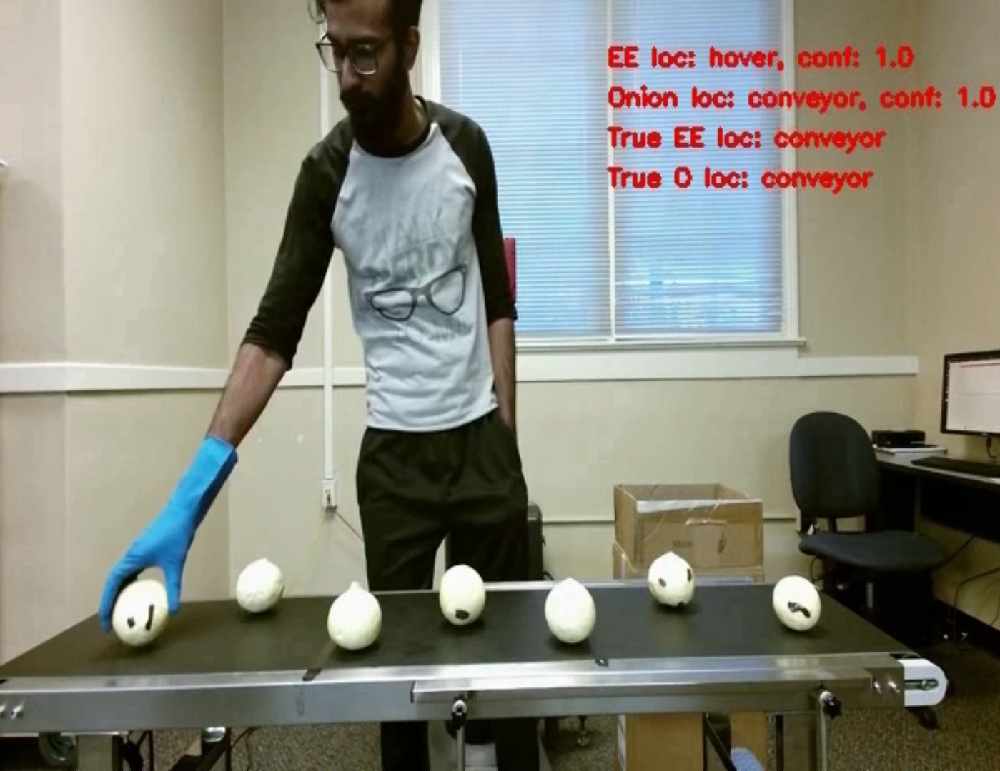}  
  \caption{}
  \label{fig:sanet_pick}
\end{subfigure}
\begin{subfigure}{.225\textwidth}
  \includegraphics[width= 1\linewidth]{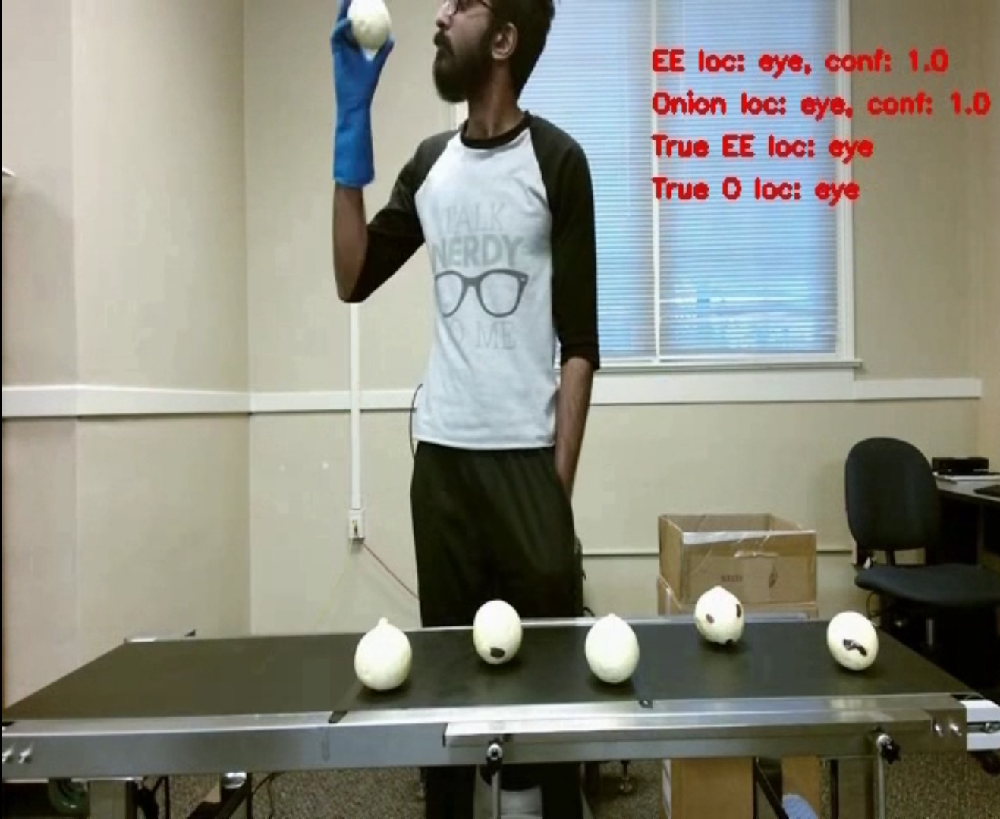} 
  \caption{}
  \label{fig:sanet_inspect}
\end{subfigure}
\begin{subfigure}{.225\textwidth}
  \includegraphics[width=1\linewidth]{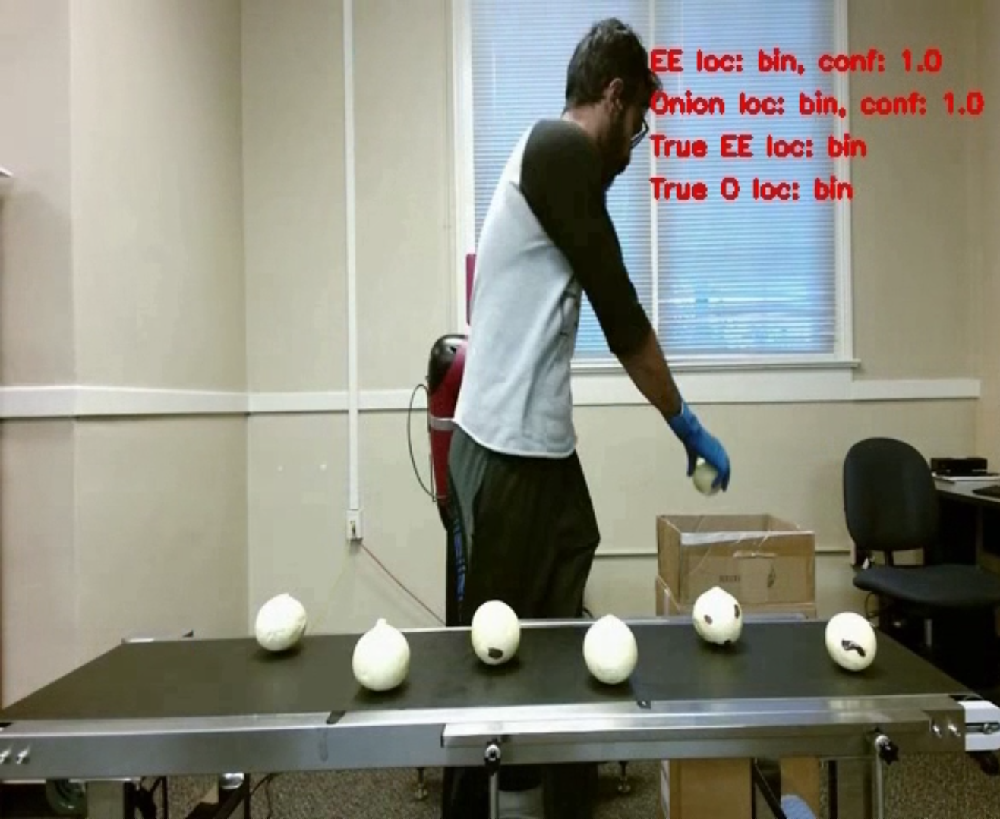}    \caption{}
  \label{fig:sanet_place2}
\end{subfigure}
\begin{subfigure}{.24\textwidth}
  \includegraphics[width=1\linewidth]{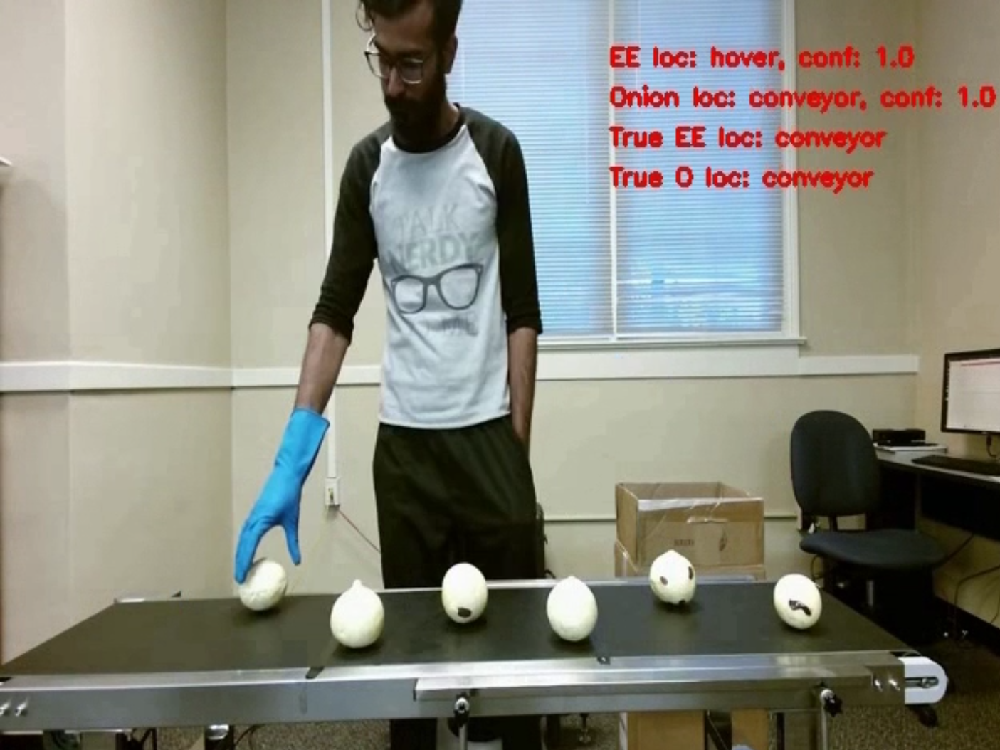}    \caption{}
  \label{fig:sanet_claim}
\end{subfigure}
\begin{subfigure}{.25\textwidth}
  \includegraphics[width= 1\linewidth]{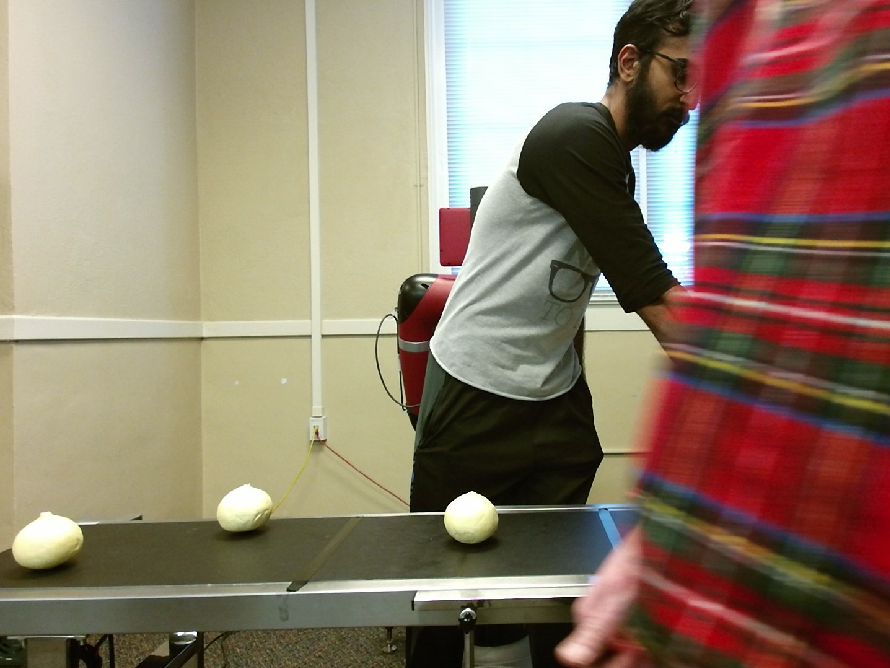}  
  \caption{}
  \label{fig:sanet_occl}
\end{subfigure}
\caption{\small (a--d) These frames show a human picking an onion, inspecting it, placing it after making a decision, and choosing the next onion in sequence. The red text appearing on the images is the state predicted by SA-Net. (e) An example occluded frame where SA-Net is unable to make a prediction. At this point the expert could be placing the onion back on the conveyor or in the bin.}
\vspace{-0.25in}	
\label{fig:Sanet-Images}
\end{figure}

Our second domain is a use-inspired robotic line sorting where the physical cobot Sawyer (from Rethink Robotics) is tasked with sorting onions on a conveyor belt after observing a human perform the sort. Sawyer observes the human, using a Kinect v2 RGB-D camera and a trained YOLO v5~\cite{redmon2016you} model is used to detect and classify the onions as blemished or not. The depth-camera frames are quantized into appropriate state variables by  SA-Net~\cite{soans2020sa}. These state recognitions are shown in red text on the frames in Fig.~\ref{fig:Sanet-Images}. 

We model the onion-sorting domain as a MDP as follows. The factored state is captured by 3 key variables yielding a total of 144 states: {\em Onion\_location}: \{0: on conveyor, 1: hover location, 2: in front of face, 3: in bin\}; {\em EndEffector\_location}: \{0: on conveyor, 1: hover location, 2: in front of face, 3: in bin\}; and {\em Prediction}:\{0: good, 1: bad, 2: unknown\}. An example state where the onion is on the conveyor, the end-effector is in the hover location, and the prediction is unknown would be represented as (0,1,2). Prior to inspection, every onion's status is unknown. The sorter performs one of five abstract actions: {\em claim new onion}, which shifts the sorter's focus to a new onion, 
{\em Pick up the onion};  {\em Inspect after picking} the onion by rotating it and checking for blemishes; 
{\em Place onion on the conveyor} after it is picked; and {\em Place onion in the bin} after it is picked. We utilize the following six Boolean feature functions to represent the human sorter's preferences:
\begin{itemize}
    \item \textsf{Good onion placed on conveyor $(s,a)$} is 1 when a good onion  is placed back on the conveyor,
    \item \textsf{Bad onion placed on conveyor$(s,a)$} is 1 when a bad onion is placed back on the conveyor,
    \item \textsf{Good onion placed in bin$(s,a)$} is 1 when a good onion is placed in the bin,
    \item \textsf{Bad onion placed in bin$(s,a)$} is 1 when a bad onion is placed in the bin,
    \item \textsf{Claim new onion$(s,a)$} is activated when a new onion is chosen if no onion is currently in focus,
    \item \textsf{Pick if unknown$(s,a)$} is 1 if the considered onion, whose classification is unknown, is picked.
\end{itemize}

The observer noise in this domain comes from YOLO sometimes misclassifying onions due to changing lighting conditions and SA-Net incorrectly identifying the state; we estimated this empirically to be approximately 30\%. This makes the state estimation uncertain and is recorded as an observation. This forms the probabilistic observation model of the camera. Occlusions occur when another person inadvertently passes by in front of the camera during the recording and blocks a frame either partially or fully, which leaves SA-Net unable to ascertain a state value (as shown in Fig.~\ref{fig:sanet_occl}).
\begin{figure}[ht!]
\centering
\begin{subfigure}{.24\textwidth}
  \includegraphics[width= 1\linewidth]{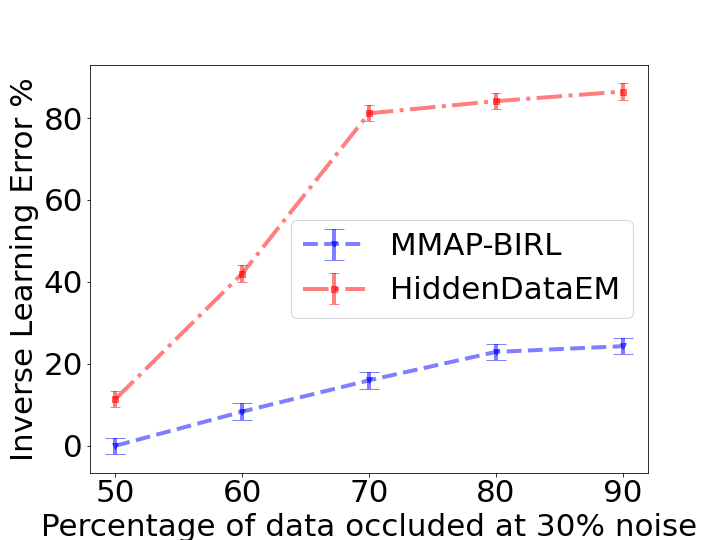} 
  \caption{}
  \label{fig:ILEvsOccl}
\end{subfigure}
\begin{subfigure}{.24\textwidth}
  \includegraphics[width= 1\linewidth]{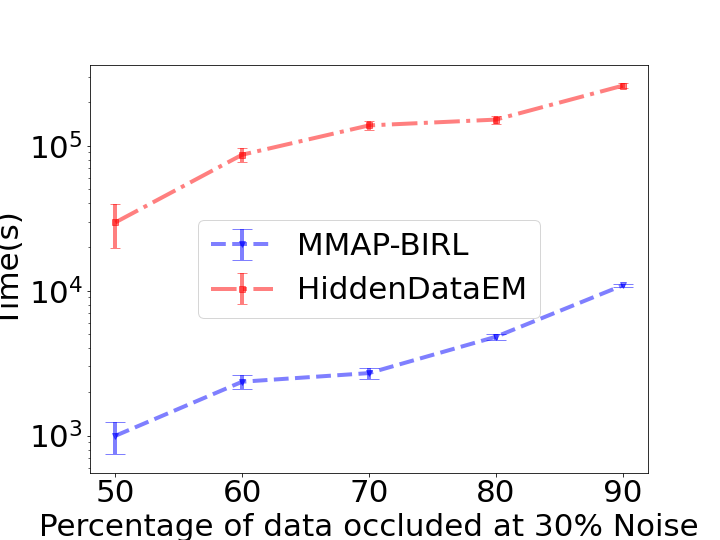} 
  \caption{}
  \label{fig:time_sorting}
\end{subfigure}
\begin{subfigure}{.5\textwidth}
  \includegraphics[width=.32\linewidth]{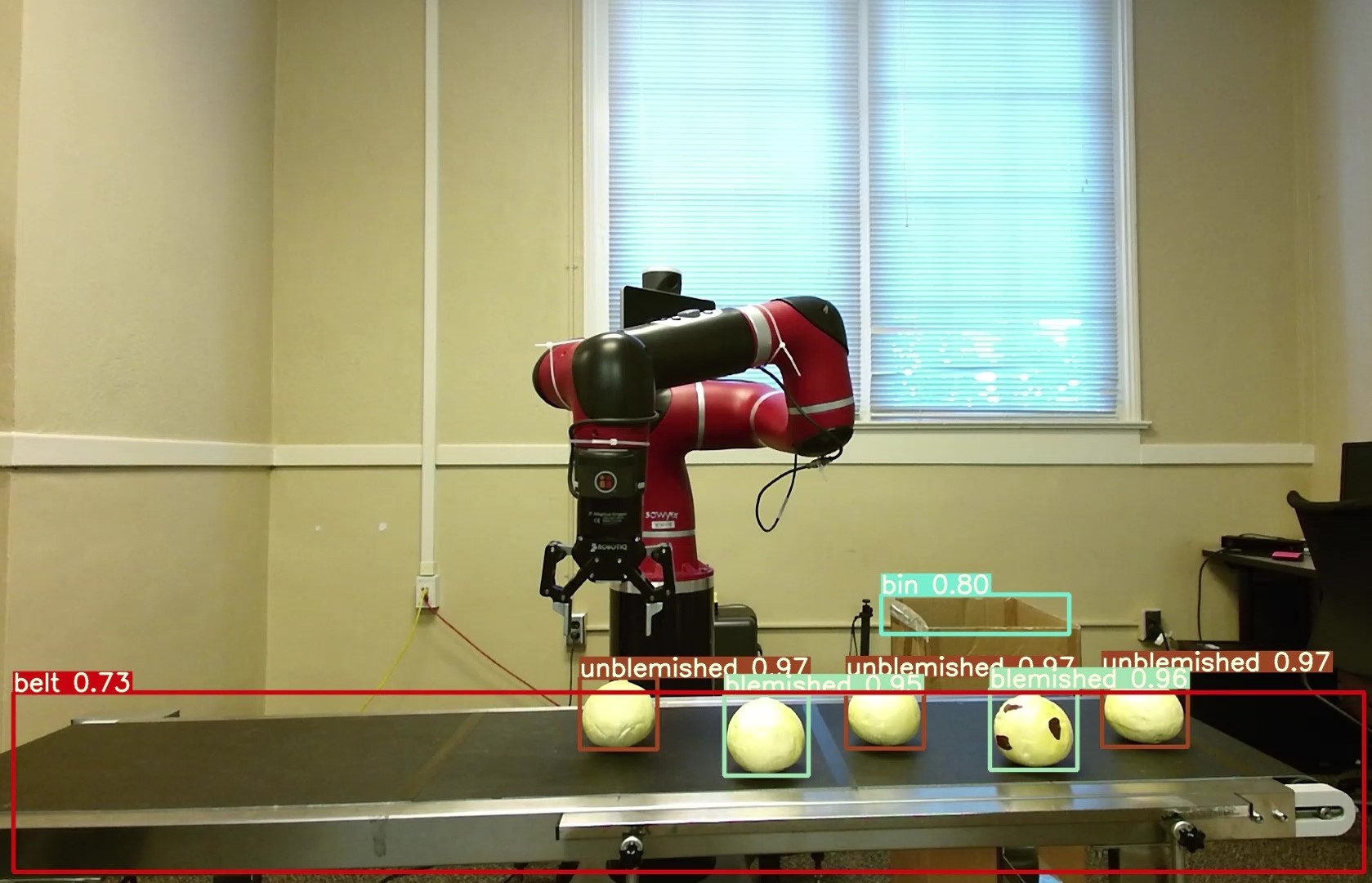} 
  \includegraphics[width=.32\linewidth]{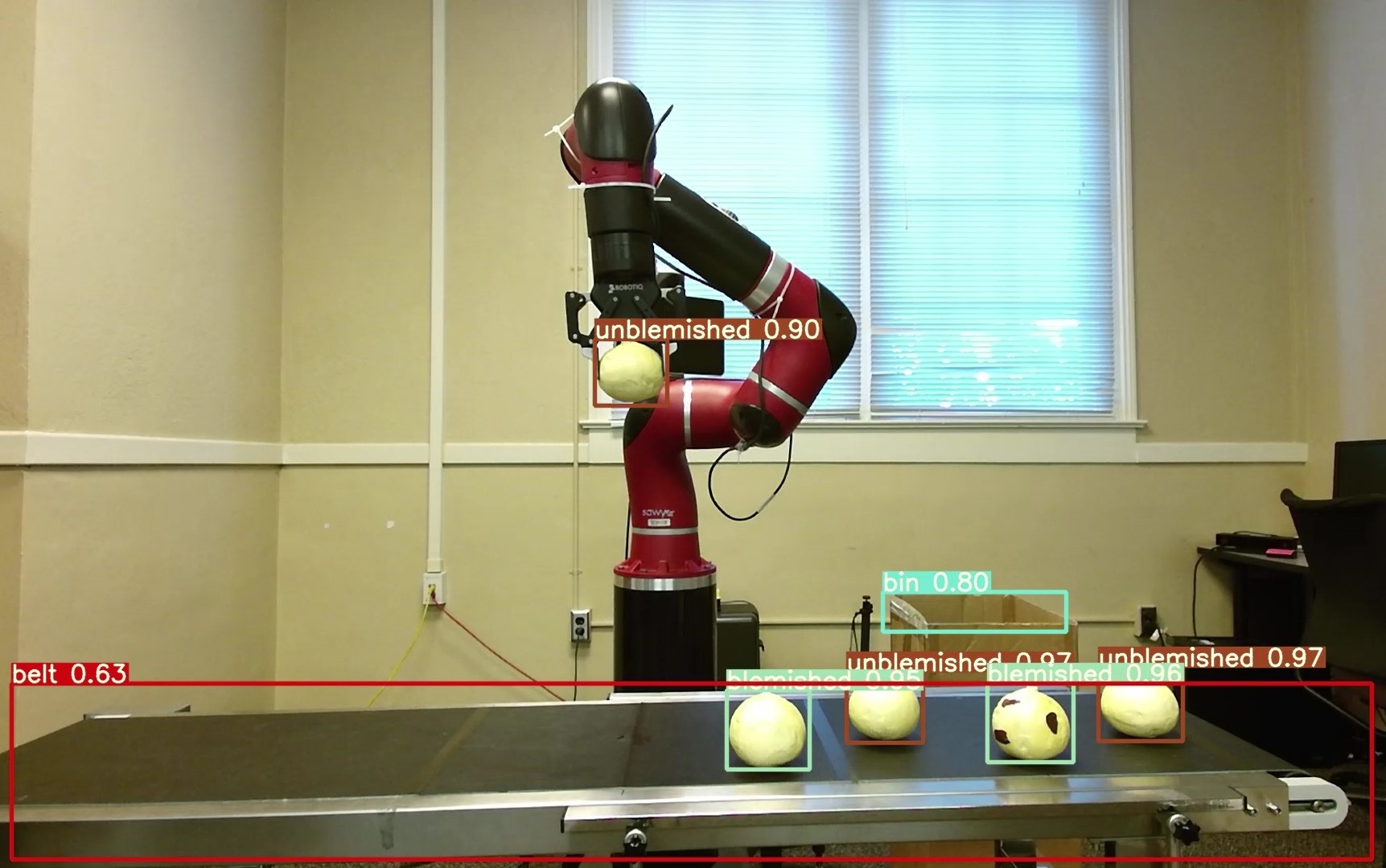} 
  \includegraphics[width=.32\linewidth]{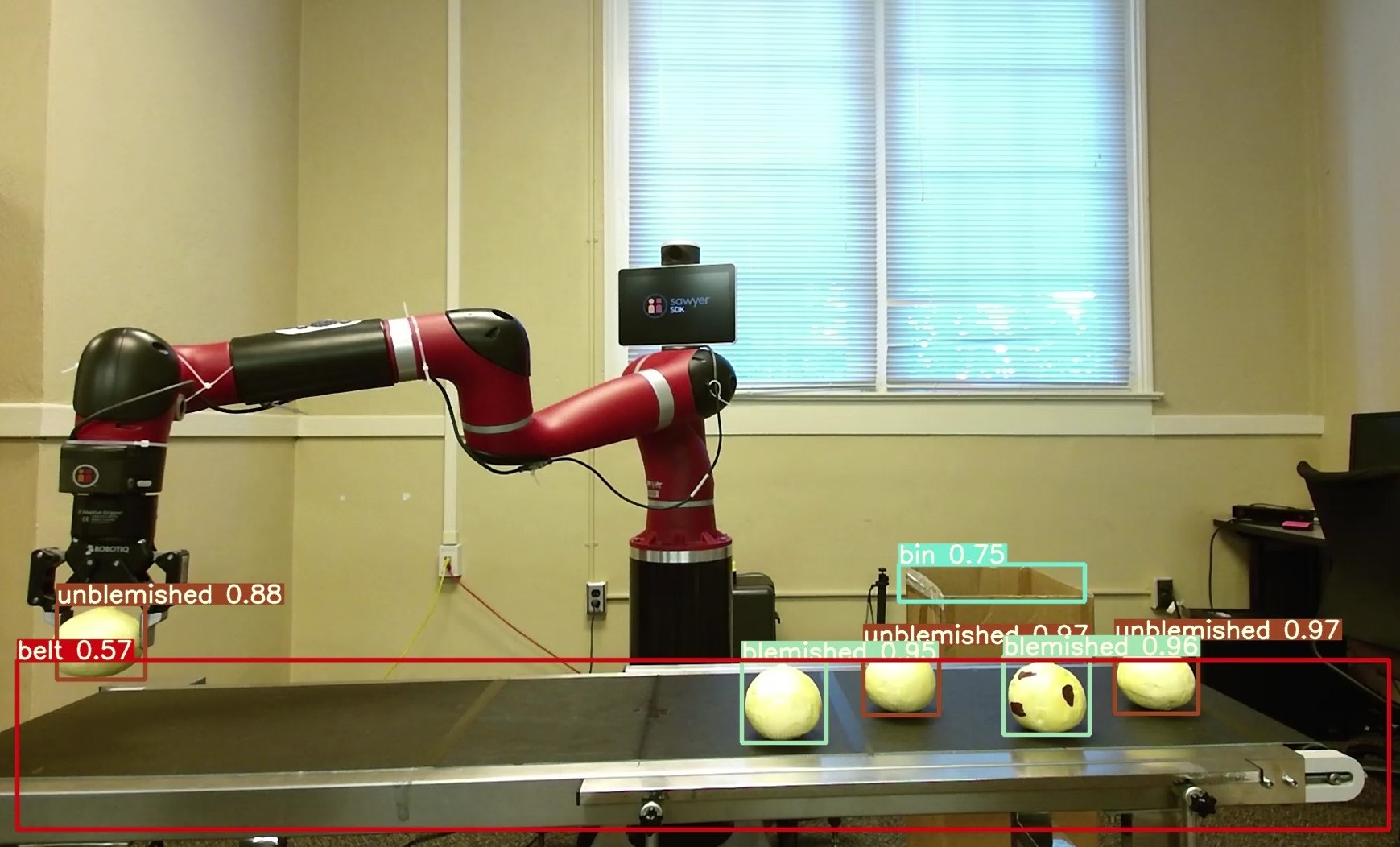} 
  \caption{}
  \label{fig:Yolo_place}
\end{subfigure}
\caption{\small (a) MMAP-BIRL exhibits much better ILE performance with occlusion \% at the estimated 30\% noise level. (b) MMAP-BIRL scales to this larger domain much better than the previous method with increasing occlusions at the same level of noise. (c) Snapshots of Sawyer sorting through the onions with bounding boxes detected in real time by YOLO v5. The run times were measured on the same computing platform as before.}
\label{fig:Occl-Yolo}
\vspace{-0.05in}	
\end{figure}
We recorded 12 trajectories from human demonstrations with an average of 4 state occlusions per trajectory. Figure~\ref{fig:ILEvsOccl} compares ILE for both MMAP-BIRL and the extended HiddenDataEM on these trajectories. Notice that MMAP-BIRL continues to show a significantly lower ILE in comparison to HiddenDataEM on this larger domain. Equally important, it does so in much less time showing more than an order of magnitude in speed up, as is evident from Fig.~\ref{fig:time_sorting}. Furthermore, the run times increase linearly in general for both methods as the rate of occlusion increases. We let Sawyer physically sort through 50 faux onions using the policies learned by both MMAP-BIRL and HiddenDataEM from the 12 recorded and processed trajectories \textbf{(see the sort video in the supplementary file)}.

\begin{table}
\begin{small}
\begin{tabular}{|c || c | c | c |} 
 \hline
 Method & (TP,FP,TN,FN) & Precision & Recall \\ [0.5ex] 
 \hline\hline
 MMAP-BIRL & (23,2,18,7) & \textbf{0.92} & \textbf{0.767} \\ 
 HiddenDataEM & (16,10,15,9) & 0.615 & 0.64\\[0.5ex] 
 \hline
\end{tabular}
\end{small}
\caption{\small Precision and recall of Sawyer physically sorting 50 onions on a conveyor using MMAP-BIRL and HiddenDataEM policies respectively.} 
\label{table:prec_recall}
\vspace{-0.35in}
\end{table}
Sawyer performs the sort by receiving the bounding boxes, on which techniques such as central orthogonal projection, and direct linear and affine transforms are used to obtain the coordinates of the onions in its 3D workspace. The robot picks up the onions and places them either in the bin or back on the conveyor after inspection. Figure~\ref{fig:Yolo_place} shows the bounding boxes detected in real-time by YOLO~\cite{redmon2016you}. Sorting performance is measured using the domain-specific metrics of precision and recall where precision = TP/(TP + FP) and recall = TP/(TP + FN). Where True Positive (TP) - blemished onions in bin, False Positive (FP) - good onions in bin, True Negative (TN) - good onions remaining on table, and False Negative (FN) - blemished onions remaining on table. From Figure~\ref{table:prec_recall}, we note that MMAP-BIRL exhibits a much better precision and recall compared to the baseline. 

\vspace{-0.05in}
\section{Related Work}
\label{sec:related}

One of the first approach to consider observer noise~\cite{shahryari2017inverse} expands the well-known maximum entropy IRL~\cite{Ziebart2008} to maximize the entropy of the joint distribution of the hidden state-action trajectories and observation sequences. A Lagrangian relaxation of this non-linear program yields gradients that can be used in the optimization. However, the gradients are hard to compute making the approach computationally unwieldy and challenging to scale. On the other hand, techniques like BIRL~\cite{zheng2014robust}, D-REX~\cite{brown2020better}, and the more recent SSRR~\cite{chen2020learning} target noisy trajectories due to the expert's failures during task performance. The former is based on the premise that noisy execution may cause the expert to sometimes follow non-policy actions. A latent variable characterizing the reliability of the action is introduced and an expectation-maximization schema in the framework of BIRL manages this noise. The latter technique (D-REX) solves the problem of automatically ranking demonstrated trajectories based on Luce-Shepard rule while SSRR uses Adversarial IRL and assumes that the demonstrator is suboptimal and that pairwise preferences over trajectories are additionally needed for IRL. However, none of these methods introduce an observation model or account for partially occluded trajectories. As the expert in our setting fully and perfectly observes its state while the learner experiences noise due to imperfect sensors, IRL methods that model the expert as a partially observable MDP (POMDP)~\cite{Choi09:Inverse} are not relevant. 

Over the past few years, there has been a steady stream of methods for inverse learning in the context of occlusion. Beginning with a method that ignores the occlusions (uses just the available data) for maximum entropy IRL~\cite{Bogert_mIRL_Int_2014}, to the HiddenDataEM, which inferred the hidden variables in actions using the expectation-maximization schema~\cite{Bogert_EM_hiddendata_fruit}, followed by ways of improving the computational tractability of the approach~\cite{Bogert17:Scaling}. As shown, MMAP-BIRL is a significant improvement over HiddenDataEM. Mai et al.~\cite{mai2019inverse} shows that forward solving the expert's MDP using a system of linear equations also allows for inferring the missing portions of the input data. But, it requires the underlying Markov chain to be non cyclic -- an assumption difficult to satisfy in practice. 
\vspace{-0.05in}
\section{Conclusion}
\label{sec:conclusion}

Motivated by the problem of learning from observing a sorting task on a line, we presented a method to generalize and improve MAP-BIRL to model and reason with both perception noise and unavoidable occlusions of portions of the data. In doing so, we show a new application of MMAP to a domain where the MAP variables are continuous, and developed a gradient-based approach to solve the MMAP inference problem. Results show that MMAP-BIRL significantly improves over the previous maximum-entropy based method for IRL under occlusions and could pave the way for facilitating future cobot deployment on factory floors.

\addtolength{\textheight}{-10cm}   




\vspace{-0.025in}
\section*{ACKNOWLEDGMENT}
\vspace{-0.025in}
This research is supported in part by NSF grant IIS-1830421 and in part by a Phase 1 grant from the Georgia Research Alliance. We thank Senthamil Aruvi for creating the 3D CAD models of the conveyor system, and Kenneth Bogert for advice and making the LME baseline available to us for evaluation. We also acknowledge valuable feedback from Saurabh Arora and Ehsan Asali, which helped improve this paper.

\bibliographystyle{IEEEtran} 
\bibliography{BIB/ICRA22,BIB/adaij16}  

\end{document}